\documentclass[letterpaper, 10 pt, conference]{ieeeconf}  

\IEEEoverridecommandlockouts                              

\usepackage{amsmath}    
\usepackage{amssymb}    

\usepackage{mathtools}
\usepackage{mathrsfs}

\usepackage{gensymb,soul}
\usepackage{graphics} 
\usepackage{subfigure}
\usepackage{booktabs}
\usepackage{epsfig} 
\usepackage{cite}
\usepackage{mathptmx} 
\usepackage{amsfonts,amssymb, amsmath}  
\usepackage[colorlinks,linkcolor=blue]{hyperref}

\usepackage{xcolor}
\usepackage{ifthen}
\usepackage{makecell}
\usepackage{algorithmicx}
\usepackage{algpseudocode}
\usepackage{threeparttable}
\usepackage[linesnumbered,ruled]{algorithm2e}

\DeclareMathAlphabet{\mathcal}{OMS}{cmsy}{m}{n}
\DeclareSymbolFont{largesymbols}{OMX}{cmex}{m}{n}

\usepackage{multirow}

\makeatletter
\let\NAT@parse\undefined
\makeatother

\title{\LARGE \bf
Interactive Navigation for Legged Manipulators \\ with Learned Arm-Pushing Controller}

\author{Zhihai Bi, Kai Chen, Chunxin Zheng, Yulin Li, Haoang Li, and Jun Ma, \textit{Senior Member, IEEE} 
 \thanks{Zhihai Bi, Kai Chen, Chunxin Zheng, and Haoang Li are with the Robotics and Autonomous Systems Thrust, The Hong Kong University of Science and Technology (Guangzhou), Guangzhou 511453, China (e-mail: zbi217@connect.hkust-gz.edu.cn). }
\thanks{Yulin Li and Jun Ma are with the Robotics and Autonomous Systems Thrust, The Hong Kong University of Science and Technology (Guangzhou), Guangzhou 511453, China, and also with the Division of Emerging Interdisciplinary Areas, The Hong Kong University of Science and Technology, Hong Kong SAR, China (e-mail: jun.ma@ust.hk). \textit{(Corresponding author: Jun Ma).}}
}%

\begin{document}

\maketitle
\thispagestyle{empty}
\pagestyle{empty}

\begin{abstract}
Interactive navigation is crucial in scenarios where proactively interacting with objects can yield shorter paths, thus significantly improving traversal efficiency. Existing methods primarily focus on using the robot body to relocate obstacles during navigation. However, they prove ineffective in narrow or constrained spaces where the robot's dimensions restrict its manipulation capabilities. 
This paper introduces a novel interactive navigation framework for legged manipulators, featuring an active arm-pushing mechanism that enables the robot to reposition movable obstacles in space-constrained environments. To this end, we develop a reinforcement learning-based arm-pushing controller with a two-stage reward strategy for object manipulation. Specifically, this strategy first directs the manipulator to a designated pushing zone to achieve a kinematically feasible contact configuration. Then, the end effector is guided to maintain its position at appropriate contact points for stable object displacement while preventing toppling. The simulations validate the robustness of the arm-pushing controller, showing that the two-stage reward strategy improves policy convergence and long-term performance. Real-world experiments further demonstrate the effectiveness of the proposed navigation framework, which achieves shorter paths and reduced traversal time. The open-source project
can be found at \url{https://zhihaibi.github.io/interactive-push.github.io/}.
\end{abstract}

\section{INTRODUCTION}

While mobile robots have achieved remarkable progress in navigating through static and dynamic obstacles \cite{intro2}, \cite{ou2024structured}, \cite{intro3}, their performance in environments containing movable obstacles remains limited. Effective navigation through such environments requires robots to actively interact with movable obstacles, such as relocating boxes or other regular-shaped obstacles, in order to create viable paths. This critical capability, known as interactive navigation \cite{stilman}, presents a fundamental challenge in enabling robots to effectively interact with movable obstacles to reshape the environment, thereby creating feasible or more efficient paths.

\begin{figure}[t]	
	\centering
	\includegraphics[width=0.85\linewidth]{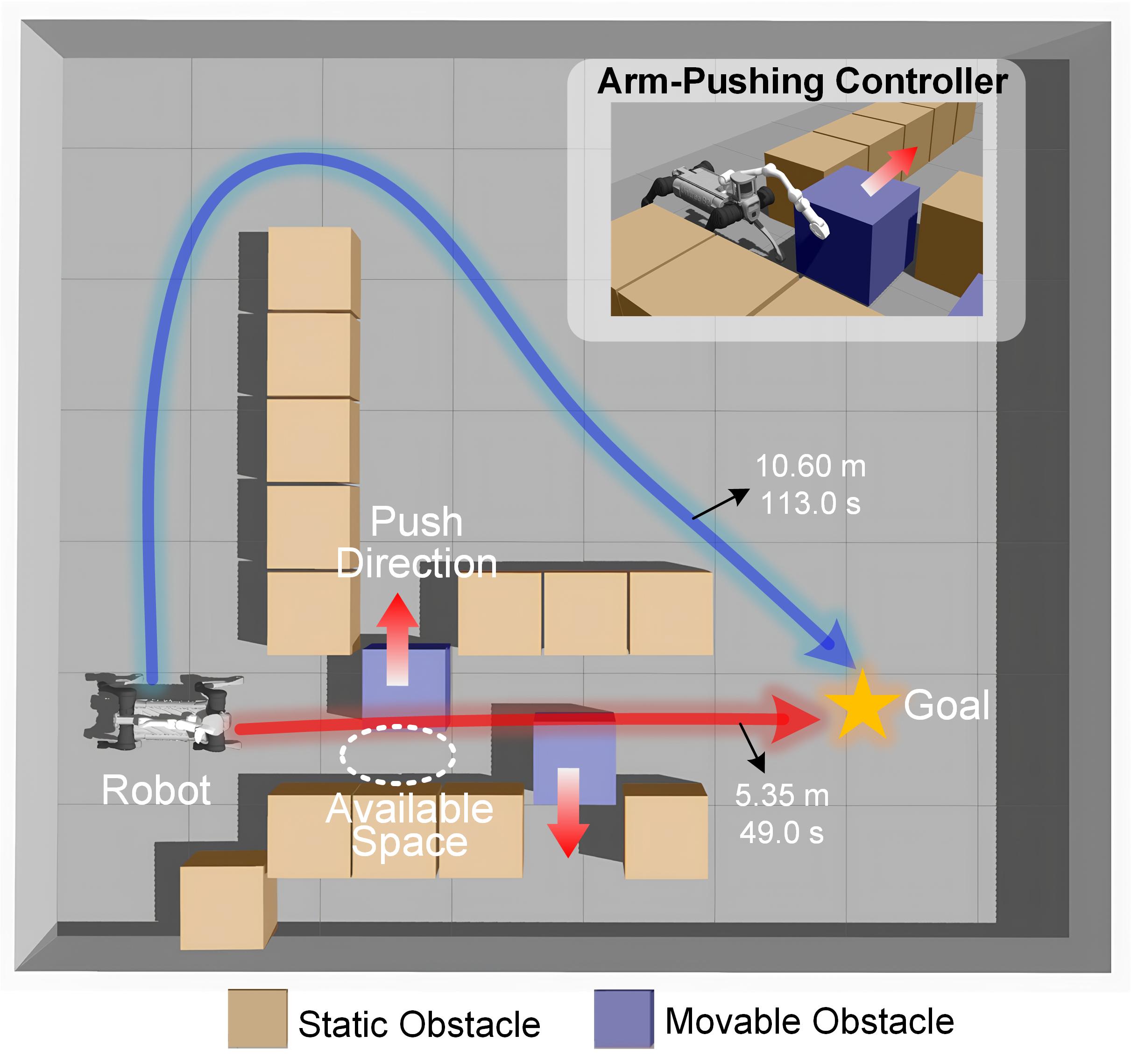}
	\setlength{\abovecaptionskip}{-0pt} 
	\caption{Illustration of interactive navigation with a learning-based arm-pushing controller in narrow spaces. 
   On the top-right of the figure, it illustrates the legged manipulator employing the arm-pushing controller to relocate the movable obstacle.
    The blue curve represents the path generated by a traditional collision-free navigation method. The {red} curve demonstrates our arm-pushing controller, which actively interacts with obstacles by pushing them to target positions, actively interacting with obstacles (e.g., pushing them aside) to create a shorter path. This enables faster goal-reaching in cluttered environments.
    }
	\label{fig:1}
\end{figure}

In current research on interactive navigation for mobile robots, some studies focus on traversability estimation \cite{schoch}, where movable obstacles are integrated into planning but without active obstacle manipulation. Other approaches explore robot-obstacle interaction, either by utilizing the robot’s body to engage with obstacles \cite{yao} or by employing a fixed-pose manipulator to assist with obstacle interaction \cite{dai}. Once contact is established, these methods typically rely on the robot body to push or manipulate obstacles. However, they require sufficient space for the robot to establish effective contact. As illustrated in Fig. \ref{fig:1}, in confined environments where the available space is smaller than the robot's body, these approaches struggle to clear obstacles efficiently, consequently hindering navigation performance. As a promising alternative, exploring an active arm-pushing mechanism with a legged manipulator provides greater flexibility. It allows the robot to clear obstacles efficiently in confined and complex environments to enhance navigation efficiency.

Essentially, pushing objects with a robotic arm is a long-standing research topic. Initially, researchers have explored analytical models and data-driven approaches, both of which have inherent limitations—analytical models require precise system dynamics \cite{nieuwenhuisen2005path}, while data-driven methods rely on large, costly real-world datasets for reliable predictions \cite{bauza2017probabilistic}. Recently, reinforcement learning (RL) has emerged as a promising alternative, as it enables low-cost trial-and-error learning in simulation, allowing the system to efficiently adapt to different objects \cite{andrychowicz2020learning}. While existing RL-based approaches primarily focus on using end-effector-mounted pushers to manipulate small, lightweight objects on 2D surfaces \cite{r4}, navigation tasks often involve pushing larger objects comparable in size to the robot itself \cite{dai}. This scale difference introduces two significant challenges:
First, large-object manipulation requires 3D spatial reasoning to handle potential manipulator-object collisions and object stability concerns, such as the risk of object toppling \cite{dadiotis2025dynamic}. Second, the policy becomes highly sensitive to the robot-object spatial relationship, with increased risk of reaching infeasible configurations. For instance, when the end-effector is positioned between the object and the target, generating effective pushing forces becomes problematic. Unlike small-object scenarios where repositioning is straightforward \cite{r5}, maneuvering around large objects requires substantial movement patterns that are difficult to discover through RL exploration.

To overcome the aforementioned challenge, we develop an arm-pushing controller leveraging RL with a well-designed two-stage reward strategy. Specifically, the proposed strategy incorporates collision avoidance constraints and the spatial relationship between the object and manipulator into the reward function. It first guides the manipulator to reach a proper pushing zone that ensures a feasible contact configuration. Then, the second-stage reward is activated to ensure that the end-effector remains within the designated pushing zone while keeping pushing the objects, which can maintain a proper contact interface for stable manipulation until the target position is reached. Such a structured reward strategy accelerates policy convergence and enhances long-term performance.

The contributions are summarized as follows:
\begin{itemize}
 
    \item We propose a novel interactive navigation framework, featuring an active arm-pushing mechanism for a legged manipulator. The framework enables the robot to reposition large movable obstacles in space-constrained environments, yielding more feasible and efficient paths.
    \item  We develop an RL-based arm-pushing controller with a two-stage reward strategy for relocating objects with a relatively large size. This progressive refinement strategy optimizes action sequences and accelerates policy convergence.
    \item Through extensive simulations and real-world experiments, we validate the robustness and generalizability of the arm-pushing controller, as well as the effectiveness of the proposed navigation framework, which enables more efficient navigation in confined spaces.
\end{itemize}

\section{Related Work}
\subsection{Interactive Navigation}

Interactive navigation aims to enhance a robot’s ability to traverse structured environments with movable obstacles, particularly in cluttered and confined spaces. The early approach in \cite{schoch} focuses on environmental awareness through RGB-D sensors to compute traversability scores. While effective for passive navigation, these methods lack active obstacle manipulation, limiting their utility in scenarios where obstacles must be physically moved. To address this issue, an RL-based method \cite{yao} is proposed to enable quadruped robots to push obstacles via body contact. However, reliance on body contact severely restricts precision and adaptability, particularly in narrow environments requiring fine-grained control. In \cite{dai}, it attempts to improve manipulation by incorporating a fixed-pose robotic arm, yet their method still depends on body velocity and lacks adaptive capabilities. Recent advancements in \cite{ellis2022navigation} explore navigation among movable obstacles, but their evaluations are limited to open spaces, leaving their effectiveness in constrained environments unverified. Unlike previous methods that push obstacles with the robot body, our framework integrates an arm-pushing mechanism. This allows the legged manipulator to actively reposition obstacles in narrow environments.

\begin{table}[t]
    \centering
    \caption{NOMENCLATURE}
\begin{tabular}{cc}
    \toprule
     \textbf{Symbols}& \textbf{Descriptions} \\
     \midrule
     $\mathbf{q} \in \mathbb{R}^6$ & Joint angles of the manipulator \\
     $^A\mathbf{p} \in \mathbb{R}^3$ & Position in arm base frame (Frame A) \\
     $^E\mathbf{p} \in \mathbb{R}^3$ & Position in end-effector frame (Frame E) \\
     $\boldsymbol{\eta} \in \mathbb{R}^3$ &  Orientation consisting of yaw, pitch, and roll angles\\
     $\boldsymbol{\mathcal{B}}_{\text{static}}$ & Set of static obstacles \\
     $\boldsymbol{\mathcal{B}}_{\text{move}}$ & Set of movable obstacles \\
     $\boldsymbol{\mathcal{P}}_{\text{obj}}$ & Set of feasible  target positions for pushing\\
     $\mathcal{C}(\Pi)$ & Occupied region of the planned path \\
      $\mathcal{C}(\mathcal{B})$ & Occupied region of the obstacles \\
     $\mathcal{Q}(b)$ & Interaction region of movable obstacle $b \in \boldsymbol{\mathcal{B}}_{\text{new}}$ \\
     \bottomrule
\end{tabular}
    \label{tab:some_definition}
\end{table}

 \begin{figure*}[t]	
	\centering
	\includegraphics[width=1\linewidth]{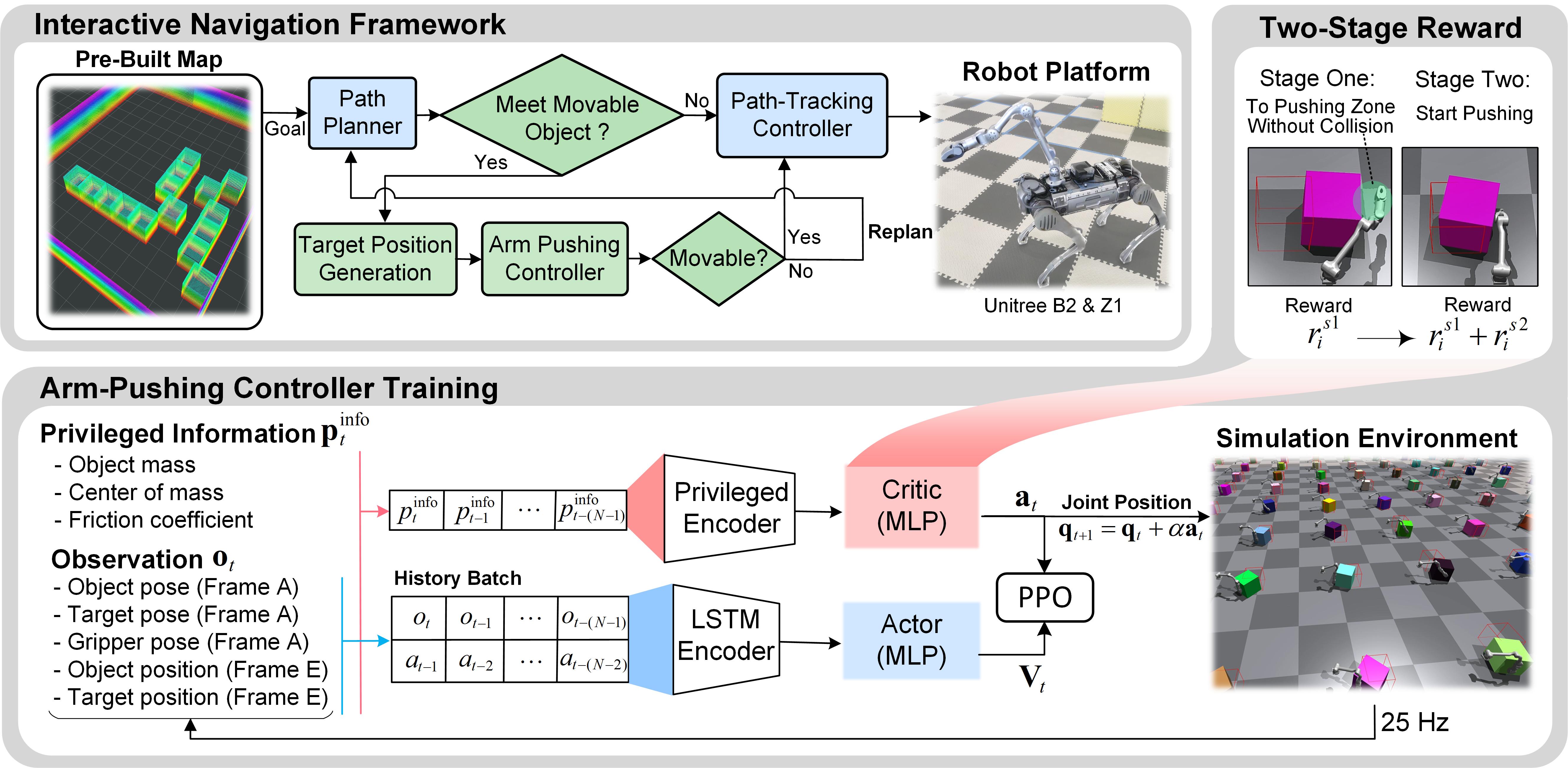}
	\caption
	{ 
    Framework of interactive navigation. 1) 
    \textbf{Interactive Navigation}: 
    Our method integrates conventional collision-free path planning, illustrated by blue boxes in the flowchart, and an arm-pushing mechanism, represented by green boxes. The pipeline includes three key stages: path planning, movable obstacle pushing, and replanning. When movable obstacles are detected, the arm-pushing controller dynamically repositions them to ensure efficient navigation.
    2) \textbf{Arm-Pushing Controller}: The pushing policy is trained to adapt to randomized physical properties (e.g., mass, friction) and object poses, ensuring robust performance across diverse scenarios. Additionally, a two-stage reward strategy is designed to accelerate convergence by separating the learning process into distinct phases.}
    \label{fig:2}
\end{figure*}

\subsection{Pushing Strategies for Object Manipulation}

Robotic object pushing is a key skill, enabling interaction with large, heavy, or irregularly shaped objects \cite{r1}. It supports essential tasks like obstacle clearing and clutter reconfiguration, making it vital for robotic systems. Early approaches to object pushing relied on analytical models \cite{nieuwenhuisen2005path}. These models assume negligible inertial effects and define motion through contact forces and friction. While they provide a strong theoretical foundation, they require precise physical parameters, such as mass and friction coefficients, which are often difficult to obtain in real-world scenarios. To address this limitation, researchers explored data-driven approaches \cite{zhou2016convex}. These methods leverage sensor feedback to learn object dynamics, either by identifying uncertain model parameters or by directly mapping actions to object behavior. Although they reduce reliance on explicit physics models, they introduce a new challenge: high dependency on large amounts of high-quality data. More recently, RL has emerged as a promising alternative \cite{r4}. RL enables robots to optimize pushing strategies through direct interaction with the environment, leveraging low-cost simulations to adapt to varying object properties without requiring exact physical parameters \cite{pp18}. However, RL-based methods often struggle with sparse rewards, particularly in task-level problems like object pushing, leading to inefficient exploration and slow policy convergence \cite{vecerik2017leveraging}. In our work, we propose a two-stage reward strategy for RL to improve training efficiency.

\section{METHODOLOGY}
In this work, we propose a novel framework for interactive navigation using the legged manipulator for achieving faster goal-reaching and reduced travel distance, with the overall navigation framework illustrated in Fig. \ref{fig:2}. 
First, we present our framework, detailing how to find a path that considers movable obstacles and determines the appropriate timing and location for pushing them (Section \ref{sec:IN}). Next, we detail our RL training process and present the two-stage reward design strategy, facilitating the deployment of a robust arm-pushing controller (Section \ref{sec:RL}). Pertinent notations used in this work are listed in Table \ref{tab:some_definition} for convenience.

\subsection{Interactive Navigation Framework}\label{sec:IN}
The navigation framework is illustrated in the upper part of Fig. \ref{fig:2}. We consider the navigation problem of a legged manipulator within a known environment, which is represented by a pre-built map \(\mathcal{M}\). The environment is composed primarily of static obstacles \(\mathcal{B}_{\text{static}}\) and movable obstacles \(\mathcal{B}_{\text{move}}\). During navigation, we assume that the robot has prior knowledge of which obstacles are movable. The robot's primary task is to navigate toward a specified goal position, while simultaneously avoiding collisions with obstacles. Additionally, the robot is equipped with a 6-DOF robotic arm, which allows for additional manipulation capabilities during navigation.

\subsubsection{\textbf{Path Planner}} 
In our framework, the global path planning process begins by generating an optimal trajectory for the robot to navigate from its initial position to the goal. This path is computed using the hybrid A* algorithm, which produces a sequence of states \(\mathcal{P}_{\text{path}} = [\mathbf{p}_0, \mathbf{p}_1, \dots, \mathbf{p}_n]\), where each state \(\mathbf{p}_i=\{x_i, y_i, z_i\}\) represents the position in the workspace. At each planning step, if the planned path overlaps with a movable obstacle, we add an extra cost $c_{t}$ based on the average time required to push the obstacle. This encourages the planner to find a time-efficient path. Then, the coarse path is refined within feasible regions to enhance navigability and continuity while ensuring smooth transitions, which is achieved by solving a quadratic programming problem: 
\begin{equation}\label{eq1}
\begin{aligned}
\min \quad & f_{\text{total}} = \lambda_1 f_s + \lambda_2 f_l \\
\text{s.t.} \quad 
& x_0 = \tilde{x}_0, \quad y_0 = \tilde{y}_0, \quad x_n = \tilde{x}_n, \quad y_n = \tilde{y}_n, \\
& x_{bi}^{-} < x_i < x_{bi}^{+}, \quad i \in \{1, 2, \dots, n-1\}, \\
& y_{bi}^{-} < y_i < y_{bi}^{+}, \quad i \in \{1, 2, \dots, n-1\},
\end{aligned}
\end{equation}
where $(\tilde{x}_0, \tilde{y}_0)$ and $(\tilde{x}_n, \tilde{y}_n)$ represent the start and goal position, $(x_{bi}^{-}, y_{bi}^{-})$ and $(x_{bi}^{+}, y_{bi}^{+})$ are the coordinates of the lower left and upper right corners of the $i$th feasible area. For more details on feasible area generation, please refer to \cite{gao2018online}.
$\lambda_1$ and $\lambda_2$ are the weight
coefficients corresponding to smoothness and length cost, which are defined as:
\begin{equation}
\begin{aligned}
    f_s &= \sum_{i=0}^{n-2} \left( (x_i + x_{i+2} - 2x_{i+1})^2 + (y_i + y_{i+2} - 2y_{i+1})^2 \right), \\
    f_l &= \sum_{i=0}^{n-1} \left( (x_i - x_{i+1})^2 + (y_i - y_{i+1})^2 \right).
\end{aligned}
\end{equation}
The optimization problem \eqref{eq1} is solved using the OSQP \cite{osqp}, an off-the-shelf optimization solver. Finally, the smoothed path is sent to a PID-based path-tracking controller for accurate execution, which guides the robot to the goal. Additionally, a replanning strategy is introduced to address situations where a movable obstacle becomes immobile due to unforeseen constraints, such as being blocked by surrounding objects.

\subsubsection{\textbf{Obstacle Relocation}}
In this part, we address when and where to relocate movable obstacles during path tracking. To ensure that the robot can effectively navigate around dynamically introduced obstructions, we continuously monitor the distance between these movable obstacles and the planned path. If a movable obstacle \(b \in \mathcal{B}_{\text{move}}\) is found to obstruct the planned trajectory and the robot enters an interaction region around the obstacle, the robot's pushing controller is triggered to displace the obstacle, thereby clearing the path. The interaction region is defined as follows:
\begin{equation}
    \mathcal{Q}(b) = \left\{ \mathbf{p}_{\text{robot}} \mid \|\mathbf{p}_{\text{robot}} - \mathbf{p}_{\text{obj}}\|_2 \leq d_{\text{push}} \right\},
\end{equation}
where \(d_{\text{push}}\) represents the activation distance. The detailed description of the pushing controller's implementation can be found in Section \ref{sec:RL}.

After entering the interaction region, it is necessary to determine an appropriate target-pushing position for the movable obstacle. To achieve this, we sample candidate positions around the centroid of the obstacle within a predefined region, and verify the feasibility of each position based on its spatial constraints. The set of feasible target-pushing positions is defined as:
\begin{equation}
\mathcal{P}_{\text{obj}} = \left\{ \mathbf{p} \sim \mathcal{D}(\mathbf{p}_{\text{obj}}, r_{\text{push}}) \mid \mathbf{p} \notin \mathcal{C}(\mathcal{P}_{\text{path}}) \land \mathbf{p} \notin \mathcal{C}(\mathcal{B}_{\text{static}}) \right\},
\end{equation}
where \(\mathcal{D}(\mathbf{p}_{\text{obj}}, r_{\text{push}})\) represents a sampling distribution centered at the obstacle's current position \(\mathbf{p}_{\text{obj}}\), with a radius \(r_{\text{push}}\). The feasible positions are those that do not intersect the planned path \(\mathcal{P}_{\text{path}}\) or static obstacles \(\mathcal{B}_{\text{static}}\), ensuring that no additional conflicts arise during the pushing operation. Once the set of feasible target positions, \(\mathcal{P}_{\text{obj}}\), is identified, the robot then selects the optimal target pushing position \(\mathbf{p}_{\text{obj}}^*\) as the closest point to the obstacle’s current position, minimizing the required pushing effort. This is formalized as:
\begin{equation}
    \mathbf{p}_{\text{obj}}^* = \arg\min_{\mathbf{p} \in \mathcal{P}_{\text{obj}}} \|\mathbf{p} - \mathbf{p}_{\text{obj}}\|_2.
\end{equation}
To push the movable obstacle to the selected optimal target position \(\mathbf{p}_{\text{obj}}^*\), we consider the task as pushing objects with varying initial states to specified target positions using a manipulator, as detailed in following Section \ref{sec:RL}.

\subsection{RL-Based Arm Pushing Policy}\label{sec:RL}
Motivated by goal-conditioned RL \cite{m1}, we formulate the object pushing problem as a finite-horizon Partially Observable Markov Decision Process.
It is defined by the tuple $\langle\mathcal{S}, \mathcal{A}, \mathcal{T}, \mathcal{G}, \mathcal{R}, \mathcal{H}, \gamma, \rho_0, \rho_g \rangle$, where $s \in \mathcal{S}$ are states, $a \in \mathcal{A}$ are actions, $\mathcal{T}(s' | s,a)$ is transition dynamics, $\mathcal{G}$ is the space of goals describing the tasks, $\mathcal{R}: \mathcal{S} \times \mathcal{A} \times \mathcal{G} \to \mathbb{R}$ is the reward function defined with goals, and $\gamma $ is discount factor. $\rho_0$ and $\rho_g$ are the distributions of the initial state and the desired goal. The goal of the manipulator is to push objects to goal states $g $ via a policy $\pi$, which is learned to maximize the expected cumulative return over the goal distribution:
\begin{equation}
    J(\pi)=\mathbb{E}_{a_t\sim\pi(\cdot|s_t,g),g\sim p_g}\left[\sum_t\gamma^tr(s_t,a_t,g)\right].
\end{equation}
We develop a pushing environment in Isaac Gym and train the policy using Proximal Policy Optimization \cite{rudin2022learning}, leveraging 1,024 parallel actors for efficient data collection and policy optimization. The training details are as follows.

\begin{table}[t]
    \centering
    \caption{HYPERPARAMETER SETTINGS}
\begin{tabular}{cc}
    \toprule
     \textbf{Reward}& \textbf{Expression} \\
     \midrule
     $r_1^{s1}$ & $k_1$ \text{exp}$(-\|\mathbf{p}_{\text{ee}} - \mathbf{p}_{zone}\|_2)$\\

      $r_2^{s1}$ &  $k_2$ \text{exp} $(-\|\mathbf{a}_t - \mathbf{a}_{t-1}\|_2)$  \\
     
     $r_3^{s1}$ &  $k_3$  \\
     

       $r_4^{s1}$ & $k_4 \mathbb{I} \left( \mathbf{p}_{\text{obj,min}} \leq \mathbf{p}_{\text{ee}} \leq \mathbf{p}_{\text{obj,max}} \right)$\\

       $r_5^{s2}$ & $k_5$ \text{exp} $(- \| z_{\text{obj}} - \frac{h_{\text{obj}}}{2}\|_2)$  \\
  
     $r_6^{s2}$ &  $k_6$ \text{exp} $(-\|\mathbf{p}_{\text{obj}} - \mathbf{p}_{\text{goal}}\|_2)$  \\
     
     $r_7^{s2}$ &  $k_7$ \text{exp} $((\mathbf{p}_{\text{goal}} - \mathbf{p}_{\text{obj}})^\text{T} \mathbf{v}_{\text{obj}} - 1)$  \\

     \bottomrule
\end{tabular}
    \label{tab:reward}
\end{table}

\subsubsection{\textbf{Observations}} As the offline training process described in Fig. \ref{fig:2}, at time step $t$, the observations $\mathbf{o}_t \in \mathbb{R}^{25}$ consist of the object pose $(^A\mathbf{p}_{\text{obj}}, ^A\boldsymbol{\eta}_{\text{obj}} )$, target pose $(^A\mathbf{p}_{\text{goal}}, ^A\boldsymbol{\eta}_{\text{goal}} )$, arm end-effector pose $(^A\mathbf{p}_{\text{ee}}, ^A\boldsymbol{\eta}_{\text{ee}} )$, object position $(^E\mathbf{p}_{\text{obj}})$, target position $(^E\mathbf{p}_{\text{goal}})$. It is worth to note that all the positions in the observations represent the positions of the centers of mass (CoM). During the training, the historical sequence of observations and privileged information are both encoded utilizing an LSTM layer. This technique can implicitly estimate the object dynamics by capturing temporal dependencies and patterns in data, and mitigate the difficulty of pushing objects with different physical properties.

\subsubsection{\textbf{Two-Stage Rewards}} Pushing a box to a target location is a task-level objective that often suffers from sparse rewards, making policy convergence challenging. The previous reward strategy for pushing the small objects \cite{r4} is not suitable for our task because they are less susceptible to spatial. To address this, we propose an innovative two-stage reward strategy. In the first stage, the arm is guided to move toward the \textit{Pushing Zone} without making contact with the object. The \textit{Pushing Zone} is defined as a cylindrical region centered at $\mathbf{p}_{\text{zone}}$ with a radius of 0.2\,m and a height equal to that of the object ($h_{\text{obj}}$). The $\mathbf{p}_{\text{zone}}$ is defined by:
\begin{equation}
    \mathbf{p}_{\text{zone}} =  \mathbf{p}_{\text{obj}} - \frac{l_{\text{obj}}}{2} \frac{\mathbf{p}_{\text{goal}}-\mathbf{p}_{\text{obj}}}{\|\mathbf{p}_{\text{goal}}-\mathbf{p}_{\text{obj}}\|_2},
\end{equation}
where the $l_{\text{obj}}$ is the length of the box. Only upon entering the  \textit{Pushing Zone} is the task-level reward activated, further directing the arm to push the object toward the target location. All the reward functions are shown in Table \ref{tab:reward}, where the $\mathbb{I}(\cdot)$ is the indicator function, which returns 1 if the condition is true and 0 otherwise. In stage one, the rewards consist of following components: encouraging the end-effector to the \textit{Pushing Zone} ($r_1^{s1}$), Promoting the smooth actions ($r_2^{s1}$), reducing time costs ($r_3^{s1}$), and avoiding the end-effector from colliding with the object ($r_4^{s1}$), where the $\mathbf{p}_{\text{obj, min}}$ is the object's minimum boundary. In stage two, the task-level rewards, $r_5^{s2}$,  $r_6^{s2}$, and $r_7^{s2}$, aim to maintain stable pushing, and prevent the object from tilting, minimize the distance between the object position and the goal position, and encourage the object's movement direction to align with the goal direction. It is worth noting that the second-stage rewards are set to zero until the end-effector reaches the \textit{Pushing Zone}.

\begin{figure}[t]	
    \centering
    \includegraphics[width=0.85\linewidth]{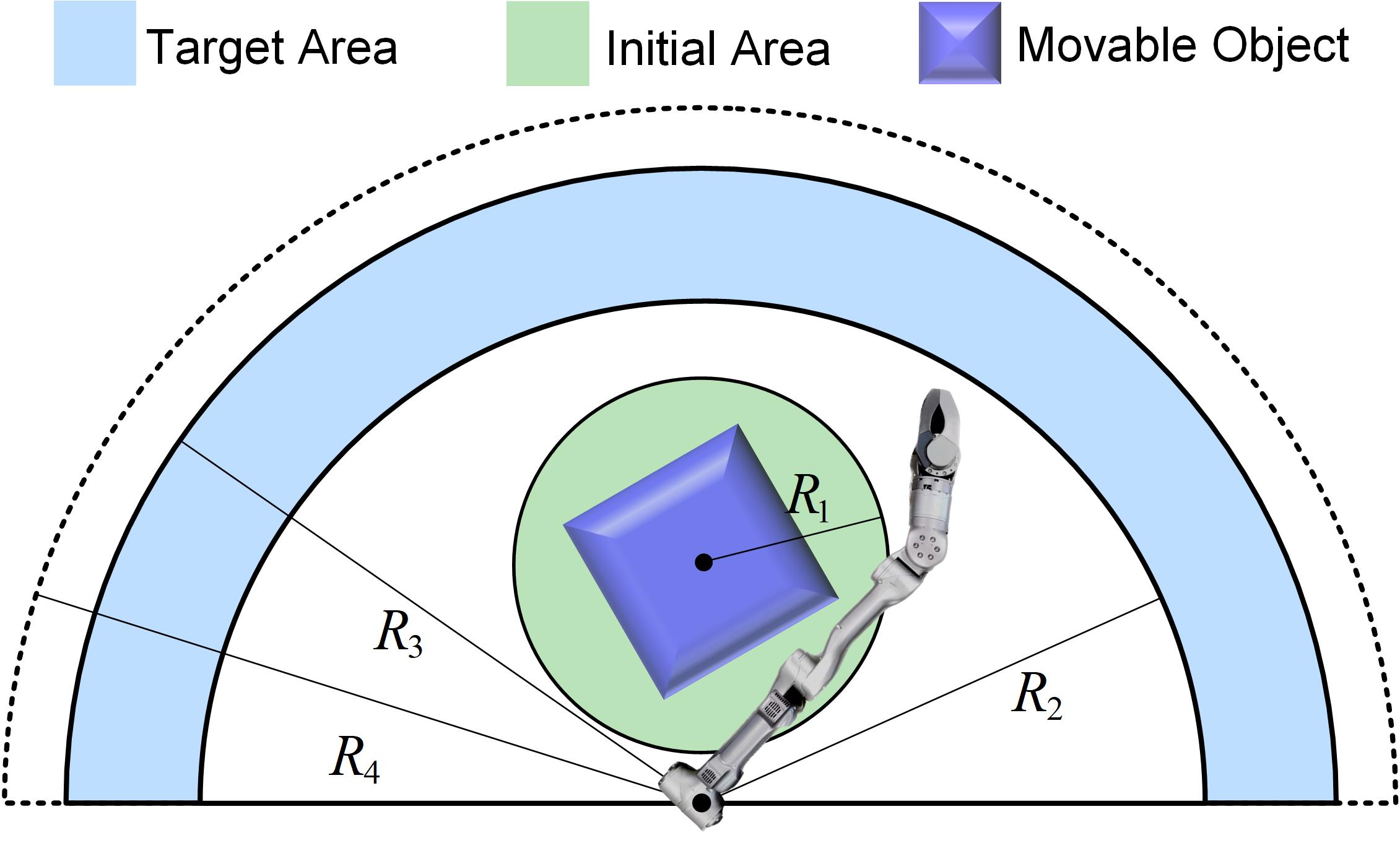}
    \setlength{\abovecaptionskip}{-0pt} 
    \caption{
    Top-down view of the training environment setting. The blue box represents the movable object. The circle with radius \( R_1 \) defines the region for the object's initial random placement, with its CoM and orientation both randomized. The light blue region, bounded by radii \( R_2 \) and \( R_3 \), represents the target area where the object's CoM is sampled. The radius \( R_4 \) defines the failure threshold; if the object's CoM exceeds the semicircular region with radius \( R_4 \), the task is considered failed, and the environment is reset.
    }
    \label{fig:3}
\end{figure}

\subsubsection{\textbf{Initial State, Goal, and Action}} 
This subsection introduces the training environment setup, including the object's initial position, and the goal position settings as shown in Fig. \ref{fig:3}, and the policy action's output. The setting of $R_i$ highly depends on the workspace of the manipulator. The goal position of the object's CoM is randomly generated within a sector-shaped area in front of the arm's base. The action vector $\mathbf{a}_t = (\Delta q_1, \Delta q_2, ..., \Delta q_6) \in \mathbb{R}^6$  consists of six arm joint position commands. The final output is the sum of the current joint angles $\mathbf{q}_t$ and the angular increments from the action output.
\begin{equation}
    \mathbf{q}_{t+1} = \mathbf{q}_t + \boldsymbol{\alpha} \mathbf{a}_t,
\end{equation}
where the $\boldsymbol{\alpha} $ is the action scale.

\subsubsection{\textbf{Curriculum Learning}}  To facilitate learning a policy for the robotic arm to push a box to a designated goal position, we adopt a curriculum learning approach. The goal position $\mathbf{p_{\text{goal}}}$ is interpolated into three intermediate goals $\mathbf{p}_1, \mathbf{p}_2, \mathbf{p}_3  $, where:
\begin{equation}
    \mathbf{p}_i = \mathbf{p}_{\text{start}} + \frac{i}{3}(\mathbf{p}_{\text{goal}} - \mathbf{p}_{\text{start}}), i=1,2,3.
\end{equation}
The policy is first trained to push the box to the intermediate goals sequentially. The transition to the next intermediate goal occurs only when the success rate of reaching the current point exceeds 90\%. 

\subsubsection{\textbf{Sim-to-Real Transfer}} For sim-to-real transfer, we implement domain randomization by randomizing various parameters related to both the arm and the environment. Specifically, the added mass of the arm is randomized within $[-1,1]$\,kg. The friction coefficient between the object and the ground is sampled from $[0.5, 1.0]$, and the object's mass is randomized within $[0.5, 3.0]$\,kg. Additionally, to enhance the robustness of the policy, Gaussian noise $\mathcal{N}$ is added to all observations.
Furthermore, for the last three joints of the arm, we utilize a high-fidelity URDF model, allowing for more realistic simulations of the contact dynamics between the arm and the object while maintaining training efficiency.

\subsubsection{\textbf{Whole-Body Pushing}}
To further enhance the manipulability, we train a whole-body controller (WBC) following \cite{liu2024visual} for the legged manipulator. This controller enables the robot to leverage its full degrees of freedom to track the end-effector pose, and this greatly expands the workspace of the manipulator. With this low-level WBC, the trained arm-pushing policy can serve as a high-level motion planner. It generates end-effector poses as inputs to the WBC, which then outputs the corresponding joint angles, enabling the robot to realize whole-body pushing. A similar hierarchical control approach is observed in \cite{xu}.

\section{Experiment}

In this section, we validate the RL-based arm-pushing controller and the interactive navigation framework through both simulation and real-world experiments. First, we evaluate the effectiveness of the learned arm-pushing controller in simulation, focusing on its adaptability and robustness in handling objects with varying physical properties, as well as the impact of the proposed two-stage reward method. Next, we conduct extensive real-world experiments to assess the practicality and efficiency of the interactive navigation framework.

\subsection{Validation of Arm-Pushing Controller}

In this subsection, we first evaluate the controller’s adaptability to diverse object properties, including size, mass, and friction. We then analyze the effectiveness of the proposed two-stage reward strategy by comparing it with two baseline approaches, highlighting its impact on learning efficiency. Finally, we observe that the controller demonstrates multi-contact interactions, dynamically adjusting contact areas to improve precision in object manipulation. 

In this work, the pushing policy operates at a frequency of 25\,Hz, with a maximum episode length of \( H = 100 \) time steps (equivalent to 10 seconds in real-time). The action scale is defined as \(\boldsymbol{\alpha} = [0.4, 0.6, 0.25, 0.25, 0.05, 0]\), and the reward function weights are set to \( k_1 = 1 \), \( k_2 = 0.1 \), \( k_3 = -0.1 \), \( k_4 = -5 \), \( k_5 = 0.2 \), \( k_6 = 1 \), and \( k_7 = 0.2 \). For the environment setup, $R_1 = 0.15$\,m, $R_2 = 0.6$\,m, $R_3 = 0.75$\,m, and $R_4 = 0.9 $\,m. The PPO hyperparameters follow the configuration in \cite{rudin2022learning}. All training is conducted on a single workstation equipped with an Intel Core i9 3.60 GHz processor and a GeForce RTX 4080 GPU.

\subsubsection{\textbf{Adaptability to Different Object Properties}} 

We evaluate the controller’s performance through sim-to-sim transfer in Gazebo, testing it on objects of varying sizes, masses, friction coefficients, and geometries. The target positions are set as (0.65\,m, 0.6\,m) or (0.65\,m, -0.6\,m) corresponding to pushing the object to the left or right, respectively. A push is deemed successful if the distance between the object's CoM and the target position is less than 0.1\,m. The success rates reported in Table \ref{tab:success_rate} are based on the statistical results of 50 repeated trials to ensure reliability. To systematically evaluate the impact of object size on the learned policy, we conduct experiments while keeping the friction coefficient fixed at 0.7 and the object mass at 1.5\,kg. We test the policy using three different box sizes, with the corresponding results presented in the first three rows of Table \ref{tab:success_rate}. The findings indicate that, despite encountering box sizes outside its training distribution, the policy retains a notable degree of adaptability.

\begin{table}[t]
    \centering
    \caption{SUCCESS RATE OF PUSHING POLICY FOR OBJECTS WITH DIFFERENT SIZES, MASSES, FRICTION COEFFICIENTS, AND SHAPES IN GAZEBO.}
\begin{tabular}{ccccc}
    \toprule
     \textbf{Object}& \makecell{\textbf{Size} \\ \textbf{[cm$^3$]}} & \makecell{\textbf{Friction} \\ \textbf{coef}} &  \makecell{\textbf{Mass} \\ \textbf{[kg]}}   & \makecell{\textbf{Success } \\ \textbf{rate [\%]}} \\
     \midrule
     Cuboid & $\textbf{60} \times \textbf{60}  \times \textbf{60}$  & 0.7 &  1.5  & 88 \\
 
     Cuboid & $\textbf{45} \times \textbf{50}  \times \textbf{50}$  & 0.7 &  1.5  & 82  \\
     
     Cuboid & $\textbf{50} \times \textbf{70}  \times \textbf{50}$  & 0.7 &  1.5  & 76 \\

    Cuboid &   50 $\times 70 \times 55$ & \textbf{0.5}  &  \textbf{1.0}  & 90 \\

    Cuboid &   $50 \times 70 \times$ 55 & \textbf{0.7}  &  \textbf{2.0}   & 92 \\
    
    Cuboid &   $50 \times 70 \times 55$ & \textbf{1.0}  &  \textbf{3.0} & 76 \\

    \textbf{Cylinder} &   $\Phi 35 \times 55$ & $0.5$ &  $1.0$ & $80$ \\
     \bottomrule
\end{tabular}
    \label{tab:success_rate}
\end{table}

Furthermore, to examine the influence of friction coefficient and object mass, we conduct additional experiments where object size is held constant while varying these two parameters. The corresponding results are provided in rows 4–6 of Table \ref{tab:success_rate}. The data suggest that the policy exhibits a certain level of robustness, consistently maintaining an overall success rate exceeding 76\% across different conditions. This shows that variations in mass and friction do not severely compromise the effectiveness of the pushing behavior. 

Lastly, to further assess the generalization capability of the policy, we test it on a cylindrical object, despite the fact that only box-shaped objects are encountered during training. As reported in the last row of Table \ref{tab:success_rate}, the policy is evaluated using a cylinder with a base diameter of 35\,cm and a height of 55\,cm, achieving a success rate of 80\%. This result further validates the adaptability of the learned policy, suggesting its potential applicability to various object properties beyond those explicitly seen during training.

\subsubsection{\textbf{Ablation Study of Two-Stage Reward Strategy}}
\begin{figure}[tbp]	
	\centering
	\includegraphics[width=0.85\linewidth]{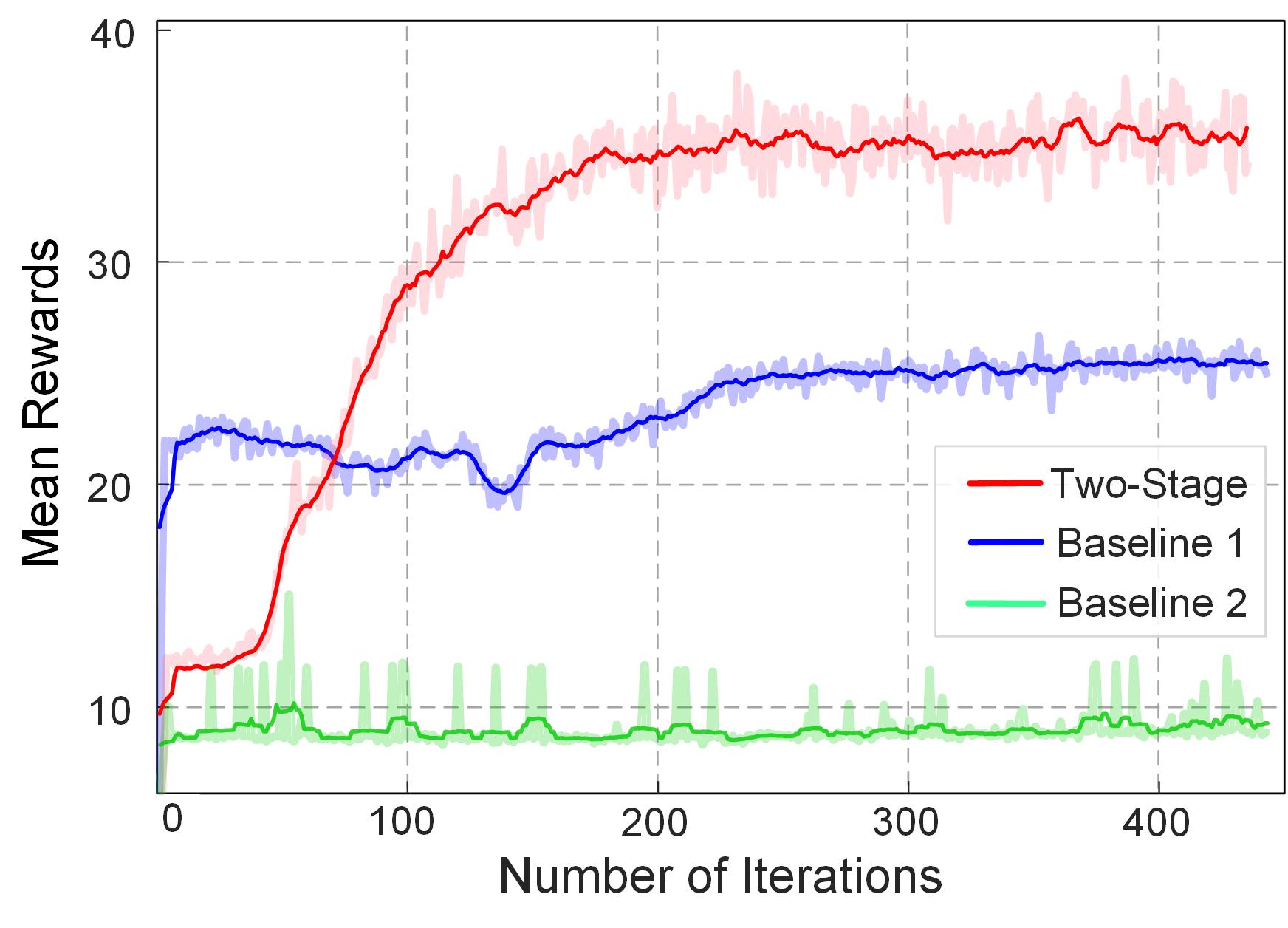}
	\setlength{\abovecaptionskip}{-0pt} 
	\caption 
	{
    Comparison of mean rewards: The blue, green, and red curves represent the mean reward trends for baseline 1, baseline 2, and the proposed two-stage reward strategy, respectively. The two-stage strategy demonstrates improved convergence and superior long-term performance compared to the other two approaches.}
    
	\label{fig:4}
\end{figure}

To evaluate the effectiveness of the two-stage reward strategy, we compare it with two baseline approaches, as illustrated in Fig. \ref{fig:4}. In baseline 1, we follow single-stage reward design principles commonly used for pushing small tabletop objects \cite{r4}. Both reward components $r^{s1}_i$ and $r^{s2}_i$ are applied from the beginning of training. So, its initial mean reward is higher compared to the two-stage strategy. However, due to the structured guidance in the two-stage approach, rewards steadily increase over time, eventually surpassing baseline 1 in both performance and convergence. In baseline 2, only the task-level reward $r^{s2}_i$ is used, leading to ineffective exploration, which renders learning more challenging. The results clearly demonstrate that the two-stage strategy not only facilitates more effective exploration but also ensures improved long-term performance compared to both baseline methods.

\begin{figure}[t]	
	\centering
	\includegraphics[width=1.0\linewidth]{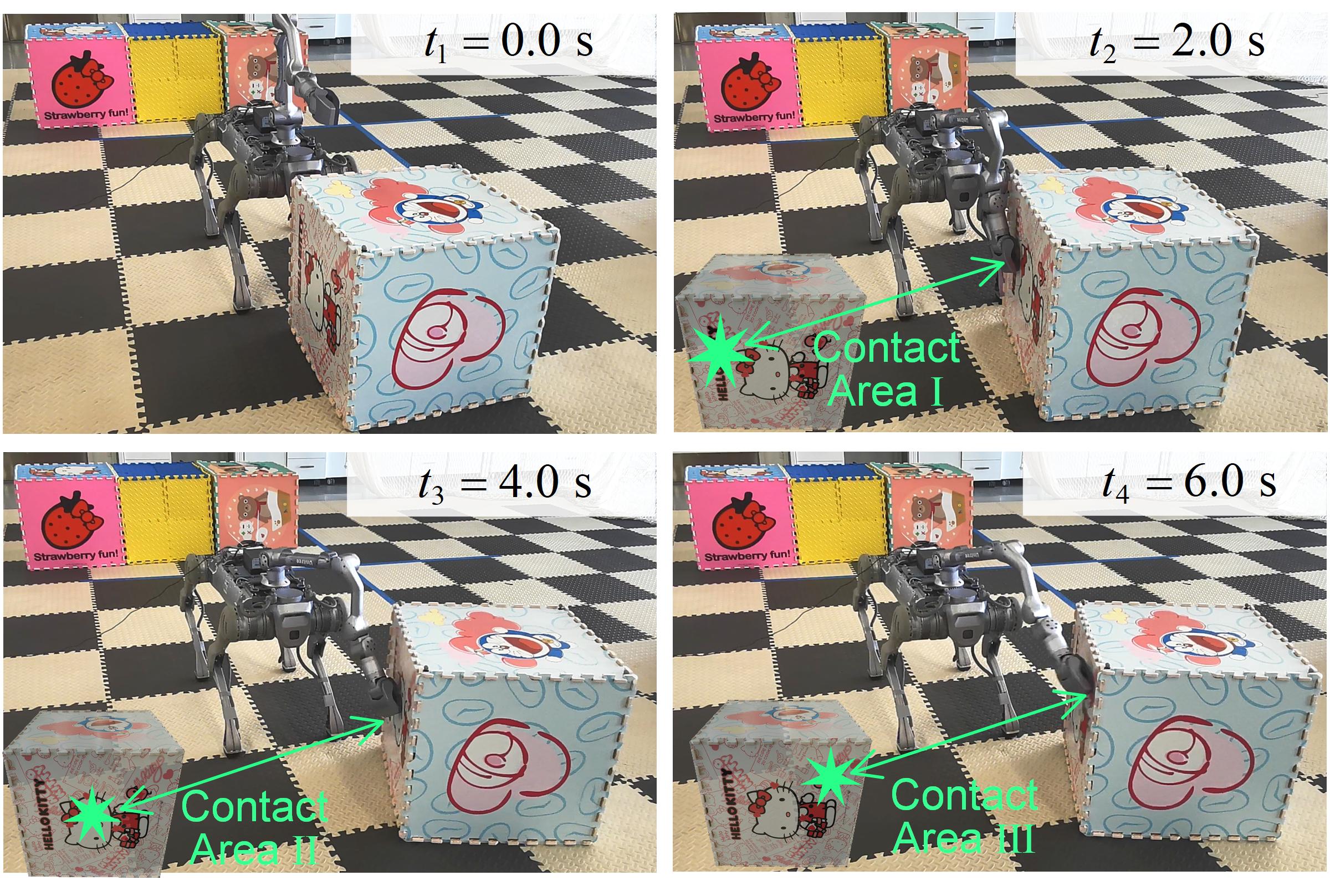}
	\setlength{\abovecaptionskip}{-0pt} 
	\caption
	{Sim-to-real transfer of the arm-pushing policy in the real robot. The sequence \( t_1 \sim t_4 \) represents a complete pushing process. Specifically, \( t_2 \) denotes the initial contact between the robot and the object, \( t_3  \) represents the second contact, and \( t_4 \) indicates the third contact. The contact areas are highlighted with green stars in the left-side view of the box.}
	\label{fig:5}
\end{figure}

\begin{figure*}[t]	
	\centering
	\includegraphics[width=1\linewidth]{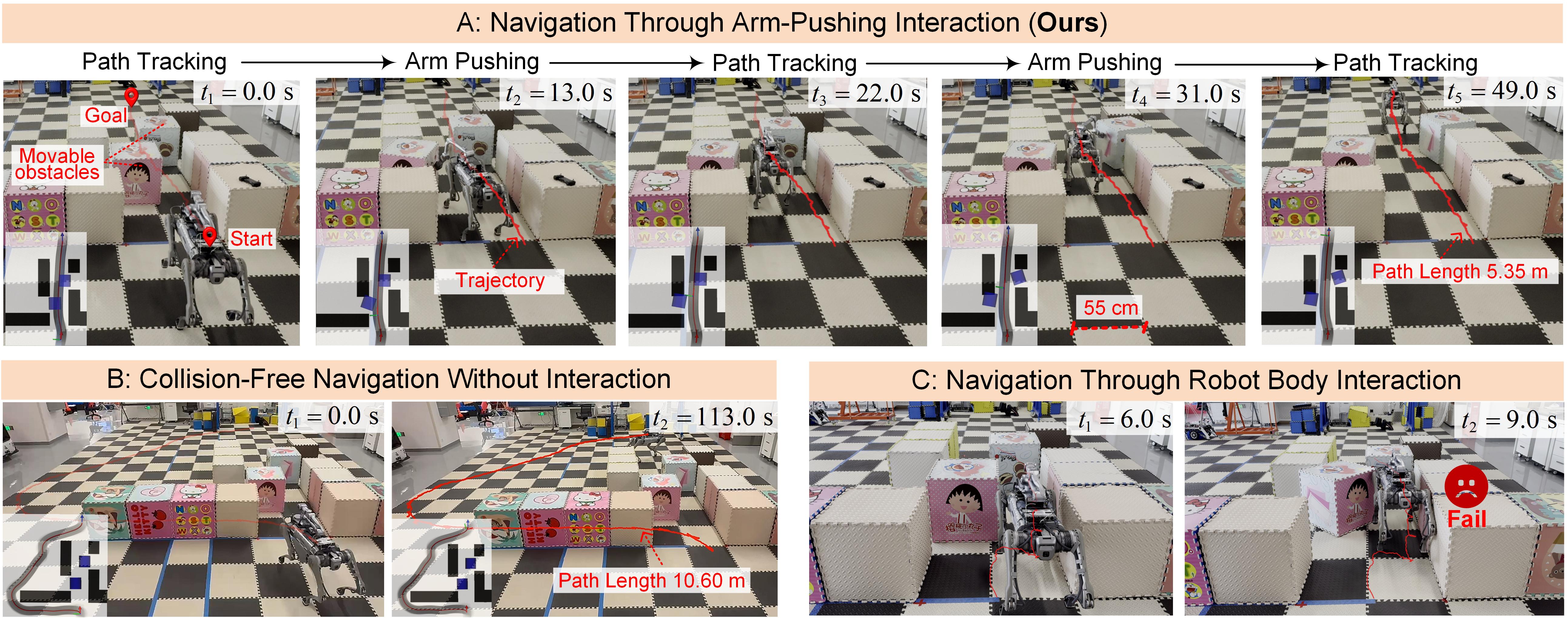}
	\caption
	{
    Comparison of navigation performance across three different methods: A) \textbf{Navigation Through Arm-Pushing Interaction}: The robot successfully navigates toward the goal by utilizing the arm-pushing controller to clear obstacles, achieving efficient and adaptive navigation. B) \textbf{Collision-Free Navigation Without Interaction}: The robot is unable to account for movable obstacles, forcing it to take detours and resulting in a longer completion time and inefficient trajectory. C) \textbf{Navigation Through Robot Body Interaction}: The robot struggles to displace obstacles effectively, resulting in task failure.}
	\label{fig:6}
\end{figure*}

\subsubsection{\textbf{Multiple Contacts with Objects}}
During the experimental deployment, we observe that the RL-based pushing controller exhibits a certain degree of flexibility and adaptability compared to a fixed action sequence controller. This is because the pushing controller is a closed-loop feedback controller that adapts its actions in real time according to the object's pose. For instance, while pushing the box, the arm makes multiple contacts with the object, gradually adjusting the contact areas to guide it toward the target location. As the light green stars shown in Fig. \ref{fig:5}, the arm interacted with the box three times— contact area $\text{I}$ on the left side ($t_2$),  contact area $\text{II}$ in the middle ($t_3$), and contact area $\text{III}$ on the upper right ($t_4$)—before the box reached its target position. To further demonstrate the robustness of the pushing controller, additional experiments under external force disturbances are provided on the project page.

\subsection{Validation of Interactive Navigation Framework} 
We deploy our framework on a Unitree B2 robot with a Unitree Z1 Pro robotic arm to evaluate its performance. An NVIDIA Jetson AGX Orin serves as the onboard computing unit for planning and control execution. As shown in Fig. \textcolor{red}{\ref{fig:6}}, we design three experimental setups: (A) Navigation through arm-pushing interaction, (B) Collision-free navigation without interaction, and (C) Navigation through the robot body interaction. For the navigation scenarios, the map measures 8\,m × 6.8\,m (consistent with Fig. \ref{fig:1}), with the planning algorithm utilizing a grid resolution of 0.1\,m × 0.1\,m. During planning, the robot is represented as a 90\,cm × 60\,cm rectangular bounding box for collision checking. The cost of object pushing $c_t = 10$, and the path planning time is about 200\,ms. In real-world experiments, the poses of the robot and movable obstacles are obtained from a motion capture system.

\subsubsection{\textbf{Navigation Through Arm-Pushing Interaction}} 
In this experiment, the arm-pushing controller is actively used to interact with and push obstacles along the robot's planned trajectory. As shown in Fig. \ref{fig:6}(A), the robot dynamically alternates between path-tracking and arm-pushing maneuvers to clear its way. The planned trajectory (marked in red) illustrates that the robot efficiently reaches its goal position while maintaining a short and smooth path. The ability to actively push obstacles allows the robot to navigate through cluttered and narrow environments, significantly improving efficiency in terms of both path length and traversal time.

\subsubsection{\textbf{Collision-Free Navigation Without Interaction}} 
In this case, the robot follows a traditional navigation strategy without utilizing the arm-pushing controller, similar to \cite{intro2}. As depicted in Fig. \ref{fig:6}(B), since the robot is unable to actively manipulate obstacles, it must find an alternative path around them. The trajectory reveals that the robot takes a significantly longer path, requiring 113.0\,s to reach the goal compared to 49.0\,s in the arm-pushing scenario—with a path length of 10.60\,m versus only 5.35\,m. The results highlight that, without interactive pushing capabilities, the robot struggles to efficiently navigate cluttered spaces, leading to increased traversal time and suboptimal path planning.

\begin{table}[t]
    \centering
    \caption{COMPARISON ON AVERAGE TRAVERSAL TIME AND PATH LENGTH}
\begin{tabular}{ccc}
    \toprule
     \textbf{Metric} & \makecell{\textbf{Traversal} \\ \textbf{Time (s)}}  &\makecell{\textbf{Path} \\ \textbf{Length (m)}}   \\
     \midrule
     Nav. W/ Arm-Pushing Interaction (\textbf{Ours}) & \textbf{58.20}   & \textbf{5.42} \\

     Collision-Free Nav. W/O Interaction  & 113.50   & 10.70  \\

     Nav. W/ Robot Body Interaction & N/A   & N/A  \\
     
     \bottomrule
\end{tabular}
    \label{tab:real_exp}
\end{table}

\subsubsection{\textbf{Navigation Through Robot Body Interaction}}
In this experiment, we evaluate robot body interaction for navigation via teleoperation, using the same interaction approach as \cite{yao}. As shown in Fig. \ref{fig:6}(C), this method results in failure—the robot becomes stuck between obstacles due to inadequate interaction control. Since the pushing force is applied through direct body contact rather than controlled arm manipulation, the robot lacks the precision needed to displace obstacles successfully. The experiment demonstrates that is less effective in structured environments, particularly in narrow spaces where precise obstacle interaction is required.

For each case, we conduct multiple repetitions in the real-world experiment to collect statistics on traversal time and path length, as presented in Table \ref{tab:real_exp}. It is important to note that navigation through robot body interaction fails to reach the goal, so its data is not included in the results. The results demonstrate the effectiveness of the proposed interactive navigation framework. To further evaluate its generalizability, we conduct additional experiments involving interactions with objects of varying shapes in diverse scenarios, which are available on the project page.

\subsubsection{\textbf{Replanning for Obstructed Movable Obstacles}}
\begin{figure}[t]	
	\centering
	\includegraphics[width=1\linewidth]{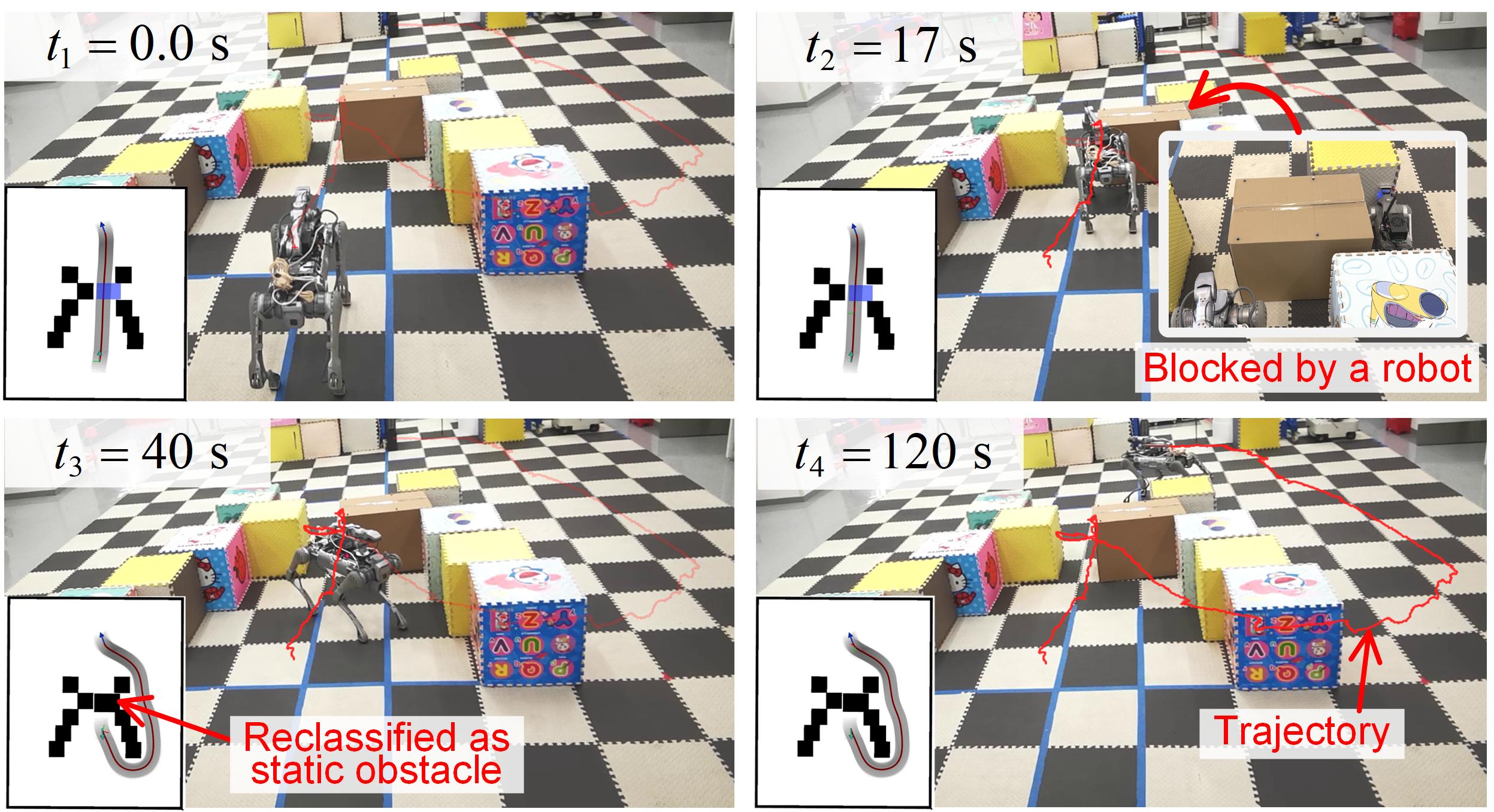}
	\setlength{\abovecaptionskip}{-0pt} 
	\caption
	{The robot replans a path when it fails to remove the obstacle. At $t_2$, the robot tries to push the obstacle. After a failed push attempt, the robot replans a path at $t_3$, and successfully reaches the goal at $t_4$.}
	\label{fig:7}
\end{figure}

We conduct an experiment to validate the effectiveness of the replanning strategy when a previously movable obstacle becomes immovable due to unforeseen factors, as illustrated in Fig. \ref{fig:7}. In this scenario, the obstacle is obstructed by another robot, rendering it unpushable. The algorithm initially attempts to push the obstacle while monitoring its displacement. When no movement is detected, the obstacle is reclassified as static, and a replanning process is triggered to compute an alternative path to the goal.

\section{CONCLUSIONS}

In this work, we propose an interactive navigation framework that enables a legged manipulator to clear obstacles actively using an arm-pushing mechanism, facilitating efficient traversal in constrained environments. The framework integrates an RL-based arm-pushing controller trained with a two-stage reward strategy, which allows the robot to effectively reposition obstacles with diverse properties. Through extensive simulations and real-world experiments, we validate the framework’s effectiveness, demonstrating its robustness in handling objects of varying mass, size, friction, and shape. The results highlight significant improvements in navigation efficiency, including reduced travel time and shorter path lengths. At this stage, our approach considers only the pushing skill within the interactive navigation framework. As part of our future work, we aim to enhance the system by incorporating more versatile manipulation capabilities to enable a broader range of interactive tasks.





\bibliographystyle{IEEEtran}
\bibliography{ref}

\begin{thebibliography}{10}
\providecommand{\url}[1]{#1}
\csname url@samestyle\endcsname
\providecommand{\newblock}{\relax}
\providecommand{\bibinfo}[2]{#2}
\providecommand{\BIBentrySTDinterwordspacing}{\spaceskip=0pt\relax}
\providecommand{\BIBentryALTinterwordstretchfactor}{4}
\providecommand{\BIBentryALTinterwordspacing}{\spaceskip=\fontdimen2\font plus
\BIBentryALTinterwordstretchfactor\fontdimen3\font minus \fontdimen4\font\relax}
\providecommand{\BIBforeignlanguage}[2]{{%
\expandafter\ifx\csname l@#1\endcsname\relax
\typeout{** WARNING: IEEEtran.bst: No hyphenation pattern has been}%
\typeout{** loaded for the language `#1'. Using the pattern for}%
\typeout{** the default language instead.}%
\else
\language=\csname l@#1\endcsname
\fi
#2}}
\providecommand{\BIBdecl}{\relax}
\BIBdecl

\bibitem{intro2}
Y.~Li, C.~Zheng, K.~Chen, Y.~Xie, X.~Tang, M.~Y. Wang, and J.~Ma, ``Collision-free trajectory optimization in cluttered environments using sums-of-squares programming,'' \emph{IEEE Robotics and Automation Letters}, vol.~9, no.~12, pp. 11\,026--11\,033, 2024.

\bibitem{ou2024structured}
Y.~Ou, J.~Fan, C.~Zhou, S.~Kang, Z.~Zhang, Z.-G. Hou, and M.~Tan, ``Structured light-based underwater collision-free navigation and dense mapping system for refined exploration in unknown dark environments,'' \emph{IEEE Transactions on Systems, Man, and Cybernetics: Systems}, vol.~55, no.~1, pp. 110--123, 2024.

\bibitem{intro3}
K.~Chen, H.~Liu, Y.~Li, J.~Duan, L.~Zhu, and J.~Ma, ``Robot navigation in unknown and cluttered workspace with dynamical system modulation in starshaped roadmap,'' in \emph{Proceedings of IEEE International Conference on Robotics and Automation}, 2024.

\bibitem{stilman}
M.~Stilman and J.~J. Kuffner, ``Navigation among movable obstacles: Real-time reasoning in complex environments,'' \emph{International Journal of Humanoid Robotics}, vol.~2, no.~04, pp. 479--503, 2005.

\bibitem{schoch}
P.~Schoch, F.~Yang, Y.~Ma, S.~Leutenegger, M.~Hutter, and Q.~Leboutet, ``In-sight: Interactive navigation through sight,'' in \emph{Proceedings of IEEE/RSJ International Conference on Intelligent Robots and Systems}, 2024, pp. 7794--7800.

\bibitem{yao}
L.~Yao, V.~Modugno, A.~M. Delfaki, Y.~Liu, D.~Stoyanov, and D.~Kanoulas, ``Local path planning among pushable objects based on reinforcement learning,'' in \emph{Proceedings of IEEE/RSJ International Conference on Intelligent Robots and Systems}, 2024, pp. 3062--3068.

\bibitem{dai}
C.~Dai, X.~Liu, K.~Sreenath, Z.~Li, and R.~Hollis, ``Interactive navigation with adaptive non-prehensile mobile manipulation,'' \emph{arXiv:2410.13418}, 2024.

\bibitem{nieuwenhuisen2005path}
D.~Nieuwenhuisen, A.~F. van~der Stappen, and M.~H. Overmars, ``Path planning for pushing a disk using compliance,'' in \emph{Proceedings of IEEE/RSJ International Conference on Intelligent Robots and Systems}, 2005, pp. 714--720.

\bibitem{bauza2017probabilistic}
M.~Bauza and A.~Rodriguez, ``A probabilistic data-driven model for planar pushing,'' in \emph{Proceedings of IEEE International Conference on Robotics and Automation}, 2017, pp. 3008--3015.

\bibitem{andrychowicz2020learning}
O.~M. Andrychowicz, B.~Baker, M.~Chociej, R.~Jozefowicz, B.~McGrew, J.~Pachocki, A.~Petron, M.~Plappert, G.~Powell, A.~Ray \emph{et~al.}, ``Learning dexterous in-hand manipulation,'' \emph{The International Journal of Robotics Research}, vol.~39, no.~1, pp. 3--20, 2020.

\bibitem{r4}
J.~D.~A. Ferrandis, J.~Moura, and S.~Vijayakumar, ``Nonprehensile planar manipulation through reinforcement learning with multimodal categorical exploration,'' in \emph{Proceedings of IEEE/RSJ International Conference on Intelligent Robots and Systems}, 2023, pp. 5606--5613.

\bibitem{dadiotis2025dynamic}
I.~Dadiotis, M.~Mittal, N.~Tsagarakis, and M.~Hutter, ``Dynamic object goal pushing with mobile manipulators through model-free constrained reinforcement learning,'' \emph{arXiv:2502.01546}, 2025.

\bibitem{r5}
X.~Sun, J.~Li, A.~V. Kovalenko, W.~Feng, and Y.~Ou, ``Integrating reinforcement learning and learning from demonstrations to learn nonprehensile manipulation,'' \emph{IEEE Transactions on Automation Science and Engineering}, vol.~20, no.~3, pp. 1735--1744, 2022.

\bibitem{ellis2022navigation}
K.~Ellis, H.~Zhang, D.~Stoyanov, and D.~Kanoulas, ``Navigation among movable obstacles with object localization using photorealistic simulation,'' in \emph{Proceedings of IEEE/RSJ International Conference on Intelligent Robots and Systems}, 2022, pp. 1711--1716.

\bibitem{r1}
F.~Ruggiero, V.~Lippiello, and B.~Siciliano, ``Nonprehensile dynamic manipulation: A survey,'' \emph{IEEE Robotics and Automation Letters}, vol.~3, no.~3, pp. 1711--1718, 2018.

\bibitem{zhou2016convex}
J.~Zhou, R.~Paolini, J.~A. Bagnell, and M.~T. Mason, ``A convex polynomial force-motion model for planar sliding: Identification and application,'' in \emph{Proceedings of IEEE International Conference on Robotics and Automation}, 2016, pp. 372--377.

\bibitem{pp18}
Y.~Geng, B.~An, H.~Geng, Y.~Chen, Y.~Yang, and H.~Dong, ``{RLAfford}: End-to-end affordance learning for robotic manipulation,'' in \emph{Proceedings of IEEE International Conference on Robotics and Automation}, 2023, pp. 5880--5886.

\bibitem{vecerik2017leveraging}
M.~Vecerik, T.~Hester, J.~Scholz, F.~Wang, O.~Pietquin, B.~Piot, N.~Heess, T.~Roth{\"o}rl, T.~Lampe, and M.~Riedmiller, ``Leveraging demonstrations for deep reinforcement learning on robotics problems with sparse rewards,'' \emph{arXiv:1707.08817}, 2017.

\bibitem{gao2018online}
F.~Gao, W.~Wu, Y.~Lin, and S.~Shen, ``Online safe trajectory generation for quadrotors using fast marching method and bernstein basis polynomial,'' in \emph{Proceedings of IEEE International Conference on Robotics and Automation}, 2018, pp. 344--351.

\bibitem{osqp}
B.~Stellato, G.~Banjac, P.~Goulart, A.~Bemporad, and S.~Boyd, ``{OSQP}: an operator splitting solver for quadratic programs,'' \emph{Mathematical Programming Computation}, vol.~12, no.~4, pp. 637--672, 2020.

\bibitem{m1}
M.~Liu, M.~Zhu, and W.~Zhang, ``Goal-conditioned reinforcement learning: Problems and solutions,'' \emph{arXiv:2201.08299}, 2022.

\bibitem{rudin2022learning}
N.~Rudin, D.~Hoeller, P.~Reist, and M.~Hutter, ``Learning to walk in minutes using massively parallel deep reinforcement learning,'' in \emph{Proceedings of Conference on Robot Learning}, 2022, pp. 91--100.

\bibitem{liu2024visual}
M.~Liu, Z.~Chen, X.~Cheng, Y.~Ji, R.-Z. Qiu, R.~Yang, and X.~Wang, ``Visual whole-body control for legged loco-manipulation,'' in \emph{Proceedings of Conference on Robot Learning}, 2024.

\bibitem{xu}
G.~Pan, Q.~Ben, Z.~Yuan, G.~Jiang, Y.~Ji, S.~Li, J.~Pang, H.~Liu, and H.~Xu, ``{RoboDuet}: Learning a cooperative policy for whole-body legged loco-manipulation,'' \emph{IEEE Robotics and Automation Letters}, vol.~10, no.~5, pp. 4564--4571, 2025.

\end{thebibliography}

\end{document}